\documentclass[sigconf,nonacm]{acmart}
\usepackage{hyphenat}

\usepackage{amssymb}
\usepackage{enumitem}
\graphicspath{{Picture/}{./}}

\title{mmRadarTwin: A Measurement-Calibrated Signal-Level Digital Twin
Platform for Indoor mmWave Radar}

\author{Jianyi Zhou}
\affiliation{%
  \institution{The University of Sydney}
  \city{Sydney}
  \state{NSW}
  \country{Australia}}
\email{jianyi.zhou@sydney.edu.au}

\author{Chenghao Zhang}
\affiliation{%
  \institution{The University of Sydney}
  \city{Sydney}
  \state{NSW}
  \country{Australia}}
\email{chenghao.zhang@sydney.edu.au}

\author{Yanli Li}
\affiliation{%
  \institution{The University of Sydney}
  \city{Sydney}
  \state{NSW}
  \country{Australia}}
\email{yali4877@uni.sydney.edu.au}

\author{Dong Yuan}
\affiliation{%
  \institution{The University of Sydney}
  \city{Sydney}
  \state{NSW}
  \country{Australia}}
\email{dong.yuan@sydney.edu.au}

\renewcommand{\shortauthors}{Zhou et al.}

\ccsdesc[500]{Computer systems organization~Sensor networks}
\ccsdesc[500]{Computer systems organization~Embedded and cyber-physical systems}
\ccsdesc[300]{Computing methodologies~Modeling and simulation}

\keywords{mmWave radar sensing, sensor systems, digital twins, signal-level simulation, Unreal Engine}

\begin{document}

\begin{abstract}
Indoor mmWave radar perception is difficult to reproduce because measured
range-angle responses depend on scene geometry, material response,
multipath, hardware conventions, and signal processing. Existing
ray-tracing and digital-twin tools often expose rendering, channel, or
path-level quantities, while radar sensing requires complex signal
products that can be processed and compared in the same domain as real
FMCW measurements. We present mmRadarTwin, a signal-level and
path-attributed digital-twin platform for indoor mmWave radar.
mmRadarTwin links a real radar measurement branch with an Unreal Engine
scene-simulation branch through a shared receive-channel and range-angle
processing interface. The simulator writes complex multi-channel receive
grids and exports per-path contribution records that identify the actor,
material tag, propagation event, and output-bin support of each simulated
return. We evaluate mmRadarTwin in an office deployment using a commodity
monostatic mmWave radar and mobile scene-capture hardware. Across 154 measured poses spanning 22 radar locations, the current physics-only path-basis simulator recalls 70.8\% of measurement-active geometry-supported response regions in the central usable field of view while exposing residuals caused by weak or missing path support, shifted responses, unsupported anchors, and missing physical mechanisms. Rather than claiming complete radar-map reconstruction or cross-room generalization, mmRadarTwin establishes a practical systems workflow for constructing, comparing, and diagnosing indoor radar digital twins.
\end{abstract}

\maketitle

\section{Introduction}
\label{sec:intro}

Millimeter-wave (mmWave) radar is attractive for indoor perception
because it can operate under poor lighting and partial occlusion while
avoiding some privacy concerns of camera-based sensing. These properties
make mmWave radar useful for indoor mapping, localization, and
human-state monitoring~\cite{Kong2025mmWaveSurvey,Sie2024Radarize}. However, indoor mmWave measurements are difficult to
reproduce. The measured response depends on multipath propagation,
material-dependent scattering, hardware non-idealities, and small-object
clutter. Nearby radar poses can therefore produce different range-angle
responses, making perception models site-sensitive and making
large-scale repeatable data collection expensive~\cite{Borts2024RadarFields}.

\begin{figure}[t]
  \centering
  \includegraphics[
    width=0.98\columnwidth
  ]{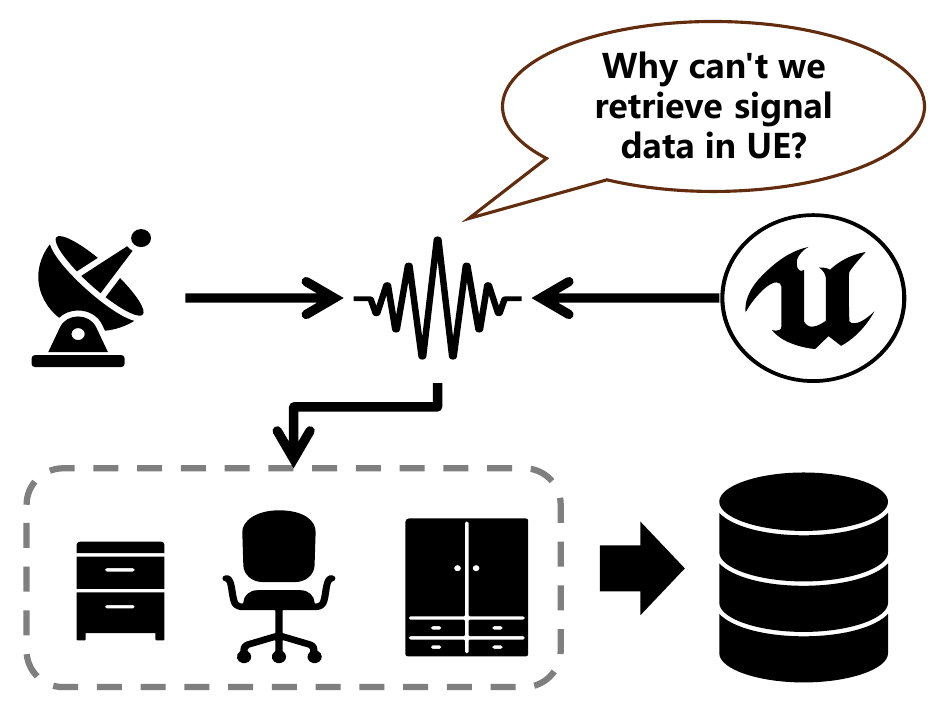}
  \caption{Motivation for mmRadarTwin. Standard game-engine scenes provide
  controllable indoor geometry and objects, but do not directly provide
  mmWave radar signal data that can be compared with real measurements.
  mmRadarTwin links radar measurements with Unreal Engine scenes and
  objects, enabling signal-level simulated radar data for comparison,
  calibration, and controlled data expansion.}
\end{figure}

A natural way to reduce this site coupling is to build a digital twin
that connects the controllable geometry of a game-engine scene with the
signal products observed by a radar. The challenge is that a game engine
does not natively emit radar measurements: it provides meshes,
materials, object poses, and collision queries, while radar sensing
requires complex receive-channel signals that pass through the same
range-angle processing chain as the real device. Most existing wireless
digital-twin pipelines target link-level quantities, such as channel
impulse responses, path-loss maps, or CSI tensors, because their target
tasks are communication-oriented rather than radar
sensing-oriented~\cite{hoydis2023sionnart,Zhang2023WiSegRT,
Salehi2024Multiverse}. For radar sensing, downstream modules such as
CFAR detection, target tracking, and point-cloud generation depend on
range-resolved spatial signal products: weak multipath returns,
sidelobe structure, relative peak amplitudes, and signal-scale
calibration all affect which peaks are detected and how they are
interpreted.

Recent work has begun to connect RF simulation and sensing from
complementary directions. RF Genesis combines ray tracing with
generative modeling to synthesize RF sensing data for learning-based
classifiers~\cite{chen2023rfgenesis}. Stereo-Fi co-trains a
differentiable ray tracer with a generative refinement model to
reconstruct 3D geometry from RF measurements~\cite{han2026stereofi}.
These systems show the promise of combining scene structure, physics,
and learning. mmRadarTwin targets a different systems layer: it builds a
signal-level radar digital-twin interface and residual-diagnosis workflow before
adding downstream learning or generative refinement.

Recent mmWave material-sensing work also clarifies the physical regime
in which our material library operates. Prior studies show that smooth
surfaces can follow Fresnel-like reflection behavior, while surface
roughness and vibration can alter the measured mmWave response
~\cite{sun2026polysight,liu2026mmtexora}.

Against this landscape, we take a signal-level systems design
perspective. Rather than making the simulator differentiable or coupling
it to a generative refinement model, mmRadarTwin implements a
deterministic forward simulator in Unreal Engine~5
(UE5)~\cite{epicgames2022ue5}. The simulator writes per-path complex
contributions into a multi-channel receive-channel (RC) grid. The same
fast-time and spatial FFT chain used for measured analog-to-digital
converter (ADC) samples is applied to the simulated RC grid, so the
measured and simulated branches terminate in a shared range-angle
comparison domain. Residuals in that domain are assigned to actionable
discrepancy categories, each mapped to an interpretable calibration
action or simulator-model limitation tag.

The sim-to-measurement gap is structured: a salient range-angle residual
is not necessarily a material error. It may reflect missing path
support, shifted geometry, system response, beamforming convention, or
an unmodeled propagation mechanism. Tuning materials until the simulated
map visually resembles the measurement can therefore hide the actual
cause. mmRadarTwin keeps path-level contribution records and routes each
residual to a targeted calibration action or simulator-model limitation
tag.

Our contributions can be summarized in three points:
\begin{enumerate}[leftmargin=*,topsep=2pt,itemsep=1pt]
\item \textbf{A signal-level indoor mmWave radar digital-twin interface.}
    mmRadarTwin connects a real measurement branch and a UE5 scene
    simulation branch through a shared range-angle signal-processing
    interface. The simulator emits per-path complex contributions into a
    multi-channel receive-channel grid, allowing simulated returns to be
    processed in the same domain as measured radar samples.

\item \textbf{A path-attributed residual diagnosis framework for
    sim-to-measurement comparison.} Residuals between the simulated and
    measured range-angle representations are routed to a structured set
    of discrepancy categories: missing path support, weak supported path,
    unsupported anchors, shifted responses, and missing physical
    mechanisms. These categories map residuals to narrower calibration
    actions or simulator-model limitation tags instead of treating every
    mismatch as a material-tuning problem.

\item \textbf{A reusable systems workflow for constructing and diagnosing
    indoor radar twins.} We implement mmRadarTwin using a commodity
    monostatic radar, mobile scene-capture hardware, and a reusable UE5
    office reconstruction. The deployment evaluates 154 measured poses
    spanning 22 radar locations and shows where the current physics-only
    path-basis simulator recovers dominant geometry-supported responses
    and where residuals remain.
\end{enumerate}

\section{Background and Requirements}
\label{sec:background}

\subsection{Indoor mmWave Sensing and Multipath}
\label{sec:background:mmwave}

Indoor mmWave sensing operates at millimeter-scale wavelengths, making room-scale objects large relative to the carrier wavelength while also making phase and scattering sensitive to small geometric and material changes. These parameters give mmWave radar fine-grained access to scene structure, but the same wavelength makes the signal sensitive to scene details that a camera or LiDAR can ignore. A single chirp's IF response, after dechirping, mixing, and a fast-time FFT, contains contributions from reflective or scattering surfaces along multi-bounce propagation paths within the radar's configured field of view, with each path's complex contribution observed across the virtual-array channels. A subsequent virtual-array angle transform yields a range-angle (RA) field on an implementation-defined grid.

For an indoor office with walls, cabinets, desks, monitors, chairs, and
soft furnishings, the dominant features of the RA field are the strong
specular returns from large planar reflectors (walls, cabinets,
monitors), the weaker diffuse returns from rough surfaces (carpets,
fabric), and multi-bounce interactions between them. Per-pose RA fields
vary considerably across nearby locations because each location samples
a different multipath geometry. Models trained on RA at one site degrade
at another, and large-scale data collection across sites is expensive.

\subsection{Why Scene Simulation Alone Is Not Enough}
\label{sec:background:gap}

A natural response is to build a simulator that replays the physical
generation process. Ray-tracing-based simulators have been used for
decades in radio propagation and now exist in modern, GPU-friendly,
optionally-differentiable forms~\cite{hoydis2023sionnart,Hoydis2024DiffRT}.
For radar-on-a-known scene, ray tracing can, in principle, enumerate
propagation paths that contribute to a range–angle response. In
practice, however, a scene simulator does not automatically become a
measurement-aligned radar twin. Two issues must be made explicit before
simulator–measurement residuals can be interpreted.

\textbf{Amplitude-scale alignment}. The measured and simulated RA
fields are produced by different amplitude chains. The measured branch
includes transmit power, antenna response, receiver gain, ADC scaling,
and downstream signal processing. The simulated branch is set by the
propagation model, per-material radar response parameters, and the
simulated noise floor. Thus, direct per-pose comparison requires an explicit
calibration-domain alignment of the two branches before residuals can be
interpreted; a single global gain constant is not sufficient.

\textbf{Complex-channel convention alignment}. The measured radar
branch and the simulator branch must use the same complex convention
before their range-angle magnitudes are compared. A sign or conjugation
difference can be invisible in magnitude-only maps while changing the
phase slope across the virtual-array channels. Detecting and correcting
such convention drift requires access to complex receive-channel data,
not just the final heatmaps.

\subsection{Platform Requirements}
\label{sec:background:requirements}

The challenges above set five requirements for an indoor radar digital
twin platform aimed at sensing rather than channel modeling. We state
these requirements before introducing the platform so that each later design
choice can be checked against a stated need.

\noindent\textbf{R1: shared signal domain.} Measured radar samples and
simulated radar samples must terminate in the same range-angle
representation through the same processing chain. Without this, a
per-cell comparison is confounded by processing-domain asymmetry, and a
discrepancy cannot be cleanly assigned to the simulator, the measurement
branch, or a convention difference between the two branches.

\noindent\textbf{R2: scene-grounded ray and path simulation.} The
simulator must operate over a 3D scene with per-actor material
assignments and coherent ray/path contributions. Without scene
grounding, a discrepancy at a given location in the radar response
cannot be traced back to a specific surface, object, or material
assignment, and the calibration workflow loses its ability to recommend
a scene-level repair instead of a generic parameter tweak.

\noindent\textbf{R3: path-level contribution attribution.} For every
simulated return that enters the shared comparison domain, the simulator
must expose a per-path complex contribution record that identifies the
actor, material assignment, propagation event, and range-angle cell
associated with the return. Without this record, the workflow has only
aggregate cell values to inspect, and discrepancies can only be guessed
at rather than explained.

\noindent\textbf{R4: single-radar reusable deployment.} The platform
should support a commodity single-radar deployment and reuse the same
scene-authoring workflow, signal-processing interface, and per-path
contribution primitive when rebuilt for a new indoor environment. This
is a systems constraint on deployment and rebuild effort, not a claim
that the workflow is tied to one radar model. Without this constraint,
moving the twin from one environment to another requires additional
radars, hardware reconfiguration, or site-specific redesign, all of
which discourage iterative use.

\noindent\textbf{R5: explicit limitation recording.} The platform should
distinguish residuals that can be addressed by scene repair, material or
RCS update, system-response handling, pose correction, or
beamforming-convention correction from residuals that expose a missing
physical mechanism. A residual that cannot be explained by the current
path basis should remain visible as a simulator-model limitation rather
than being hidden by a generic parameter tweak. Without this requirement,
the workflow may make the simulated range-angle response look closer to
the measurement while losing the causal interpretability needed for
calibration.

\section{The mmRadarTwin Platform}
\label{sec:design}

\begin{figure*}[t]
  \centering
  \includegraphics[width=0.95\textwidth]{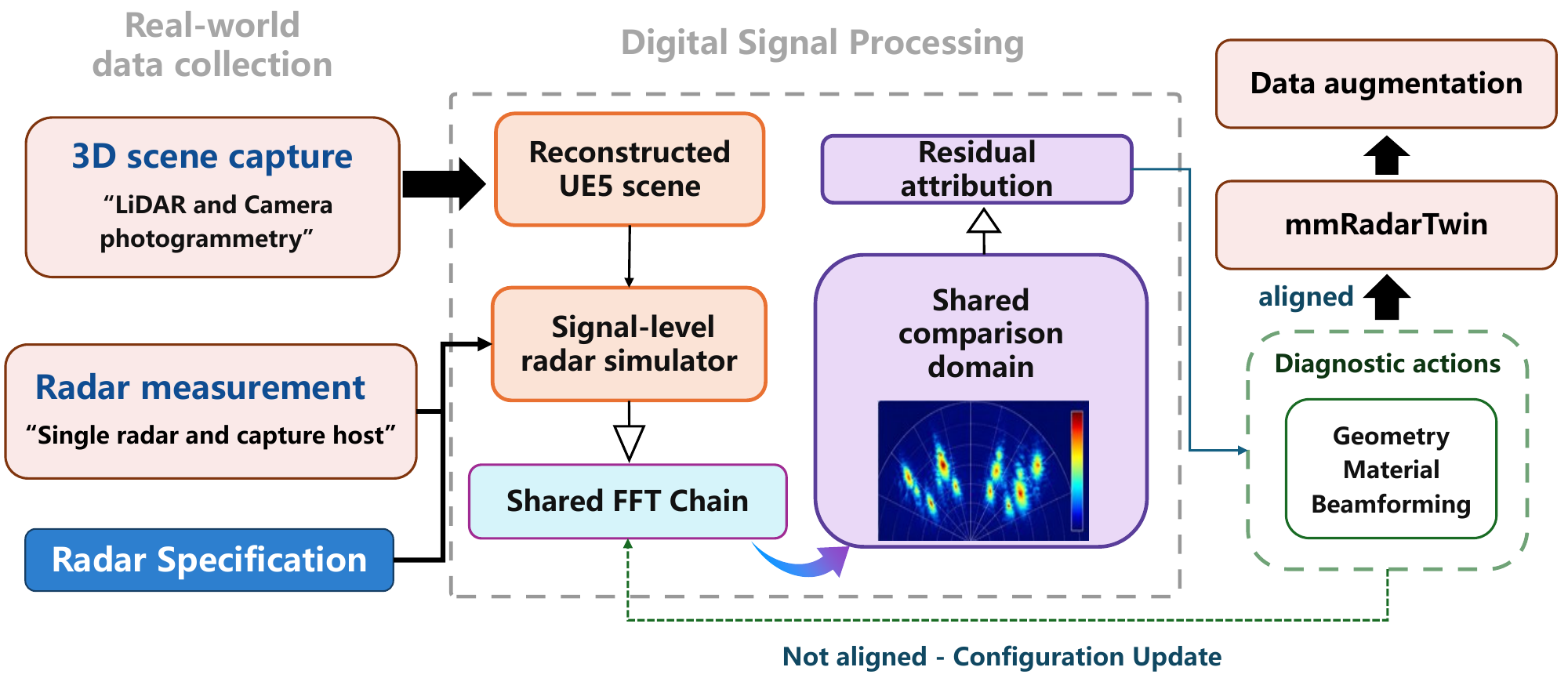}
\caption{The system overview for mmRadarTwin contains real-world capture, signal-level radar simulation, shared comparison, and calibration workflow modules. mmRadarTwin compares measured and simulated radar responses in a common range-angle domain and routes residuals to calibration actions that update the radar twin.}
\Description{Diagram showing the mmRadarTwin system overview, including modules for real-world capture, radar simulation, comparison, and calibration.}
\label{fig:system}
\end{figure*}

mmRadarTwin is a platform, not a one-shot simulation. By a platform, we
mean a fixed, reusable set of components that turns a single commodity
FMCW mmWave radar and a reconstructed indoor scene into a digital twin, whose
output can be compared, at the channel level, with real measurements.
The platform has five reusable components: a scene representation that
grounds simulated returns in named actors and materials
(Section~\ref{sec:design:scene}), a signal-level radar abstraction that
defines the receive-channel comparison interface
(Section~\ref{sec:design:radar_abstraction}), a ray and path generation
step that produces simulated propagation paths
(Section~\ref{sec:design:ray_path}), a shared signal-processing chain
that maps both measured radar samples and simulated receive-channel grids into
the same range-angle representation
(Section~\ref{sec:design:signal_processing}), and a per-path complex
contribution-record primitive that exposes attribution at the path level
(Section~\ref{sec:design:audit}).

The platform is measurement-aligned in a limited sense: the simulator's
outputs and the radar measurements are compared only after both branches
have been mapped into the same calibration domain defined by the residual
attribution framework (Section~\ref{sec:design:domain}). Thus, the
calibration domain is a contract used by the attribution framework
rather than a sixth platform component. The measured branch provides a
fixed measurement-side range-angle representation produced from raw ADC
samples through the shared signal-processing chain; this fixes the
comparison interface only and is not a simulator-versus-measurement
accuracy claim.

Rebuilding the twin for a new indoor scene under the same hardware
reuses the platform components above unchanged. The site-specific inputs
are the reconstructed actor set, the actor poses, and the per-actor
material tags.

\subsection{Scene Representation}
\label{sec:design:scene}

The platform operates over a 3D scene rather than a parametric channel
model. The scene is authored in Unreal Engine~5 and is intended to be
reusable across calibration runs, ablations, and what-if experiments.
The authoring workflow itself, including how each piece of geometry is
captured and registered to the radar's world frame, is an implementation
detail described in Section~\ref{sec:impl:scene}. What the platform
requires of any scene is a fixed structure that each piece of geometry
must expose.

Each piece of geometry is a named actor. An actor carries (i) a mesh
that resolves ray queries, (ii) a transform that places the actor in the
radar's world frame, and (iii) a material tag drawn from a radar material
library. The material tag is attached at scene-authoring time and is not
per-pose tunable; the canonical tag list and per-tag parameter values
live in the implementation chapter (Section~\ref{sec:impl:scene},
Table~\ref{tab:materials}), since they are deployment-specific rather
than platform-defining. Rebuilding the twin for a new indoor environment
under the same hardware reuses the platform unchanged: only the actor
set, per-actor transforms, and per-actor material tags are site-specific.

This per-actor structure is what makes residual attribution at the path
level possible. When the per-path complex contribution-record primitive
(Section~\ref{sec:design:audit}) emits a record, the record carries the
actor tag and material tag that produced it, so a residual at any
range-angle cell can be decomposed into the named actors responsible for
it. The scene representation answers R2 on the scene side and supplies
the input that R3 consumes.

\subsection{Signal-Level Radar Abstraction}
\label{sec:design:radar_abstraction}

The platform treats the radar as a frequency-modulated continuous-wave
(FMCW) front-end with $N_T$ transmit and $N_R$ receive antennas,
time-division multiplexed into a virtual receive array along the
azimuth dimension~\cite{Patole2017AutomotiveRadar}. Rather than committing the platform definition to a specific commercial radar module or capture board, we define the comparison interface as an $M$-channel complex receive grid, where
$M=N_TN_R$ under the virtual-array abstraction. Concrete chirp
parameters, sampling settings, and the number of virtual receive
channels are deployment-specific implementation choices.

For receive channel $m \in \{1, \ldots, M\}$ and fast-time sample
index $n \in \{0, \ldots, N_r-1\}$, the complex baseband after dechirp
is a sum over $K_p$ effective propagation paths:
\begin{equation}
  x_m[n] \;=\; \sum_{p=1}^{K_p} \alpha_{p,m} \, e^{j 2\pi f_{b,p} n T_s}
              \;+\; w_m[n],
  \label{eq:fmcw}
\end{equation}
where $\alpha_{p,m}$ is the complex coefficient of path $p$ at channel
$m$, $f_{b,p} = 2 \mu \tau_p / c$ is the beat frequency associated with
path delay $\tau_p$, $T_s$ is the ADC sampling interval, and $w_m[n]$ is
measurement noise. The coefficients $\alpha_{p,m}$ are generated by the
simulated path model, and the same range and angle processing chain is
applied to both measured and simulated receive-channel grids.

\subsection{Ray and Path Generation with Physical Mechanisms}
\label{sec:design:ray_path}

The simulated branch emits rays from each radar pose and traces them
through the reconstructed scene using shooting-and-bouncing-rays (SBR)
ray tracing~\cite{han2026stereofi}. Each valid interaction appends a
path feature to the per-pose path set
$\mathcal{P} = \{p_i\}_{i=1}^{N_p}$, carrying total propagation length
$s_p$, azimuth angle-of-arrival $\theta_p$, bounce count $b(p)$, the
actor tag, and the actor's material tag. A path terminates when it no
longer intersects scene geometry, falls below the configured power
threshold, or reaches the configured bounce limit.

\textbf{Why a ray/path abstraction.}
At the implementation wavelength reported in Section~\ref{sec:impl:hardware},
the wavelength is short relative to dominant room-scale scatterers such
as walls, cabinets, monitors, and furniture. We therefore use a
high-frequency path-basis abstraction~\cite{chen2024rfcanvas,
yun2015raytracing}: the simulator represents the dominant signal
components as surviving propagation paths, each carrying a complex
amplitude and phase, rather than solving the full wave field in the
reconstructed room. This is not a claim of full-wave accuracy. It is a
signal-level approximation whose target is the $M$-channel complex
receive grid consumed by the same range and angle FFT chain as the
measured radar. Mechanisms that are not captured by the current path
basis, such as edge diffraction, dispersive material handling, and
rough-surface scattering beyond the Rayleigh threshold, are treated as
platform extension points rather than as partially implemented physics.

We model each material using an effective complex permittivity
\begin{equation}
  \varepsilon_c \;=\; \varepsilon - j \frac{\sigma}{\omega \varepsilon_0},
  \label{eq:permittivity}
\end{equation}
where $\varepsilon$ is the permittivity, $\sigma$ the conductivity,
$\omega$ the angular frequency, and $\varepsilon_0$ the free-space
permittivity, with $\mu_r = 1$ (non-magnetic) throughout. The complex
baseband response at the receiver is a coherent superposition over
paths:
\begin{equation}
  E \;=\; \sum_{p \in \mathcal{P}} A_p \, e^{-j k_0 s_p}
          \prod_{k=1}^{N_i(p)} \Gamma_{p,k},
  \label{eq:sbr}
\end{equation}
where $k_0 = 2\pi / \lambda$ is the free-space wavenumber,
$\lambda$ is the free-space wavelength, $A_p$ encodes geometric
spreading and antenna gains, $\Gamma_{p,k}$ is the interaction
coefficient at the $k$-th hit of path $p$, and $N_i(p)$ is the number
of interactions. The interaction coefficients are parametrized by per-tag scalars. The
canonical material library, with material labels such as metal,
concrete, glass, wood, and plastic, and their per-tag values, is reported
with the scene reconstruction and material library implementation
(Table~\ref{tab:materials}). The platform itself treats the library as
deployment data, not as platform code.

Each surviving path contributes a complex receive-channel vector
$\mathbf{c}_p \in \mathbb{C}^{M}$ over the $M$ virtual channels. Its
entries share the path-level amplitude and carry channel-dependent phase
terms induced by the path angle and the virtual-array geometry. The
simulated receive grid is the coherent, phase-preserving sum of these
per-path vectors over $\mathcal{P}$; after the shared spatial FFT,
constructive and destructive interference across the virtual channels
determines the range-angle response. In the current platform, angular
discrimination is modeled through the virtual-array phase geometry and
the shared FFT chain, while measured per-element antenna patterns,
polarization, and elevation-axis beamforming remain extension points.
The scalar $A_p$ in Eq.~\ref{eq:sbr} absorbs geometric spreading and a
path-level front-end gain, not a calibrated element-pattern response.

The currently implemented mechanisms are distance attenuation, complex
Fresnel reflection in the flat-and-smooth-at-high-incidence
regime~\cite{sun2026polysight}, multi-bounce coherent superposition,
and a diffuse-backscatter response for surfaces above the Rayleigh
roughness threshold $\lambda/8$ at the implementation wavelength
~\cite{liu2026mmtexora}. Mechanisms not yet enabled, such as
uniform-theory-of-diffraction edge contributions, floor-ceiling specular
elevation paths, dispersive material handling beyond per-tag scalars,
and rough-surface scattering beyond the current threshold model, are
extension points rather than partially implemented features.

\subsection{Shared Signal Processing}
\label{sec:design:signal_processing}

The measured and simulated branches apply identical signal processing to
terminate in the same range-angle representation. A fast-time FFT over
the $N_r$ ADC samples yields complex range spectra $X_m[r]$ on each of
the $M$ virtual channels; a spatial FFT across the virtual channels then
produces the range-angle (RA) field
\begin{equation}
  S[r,a] \;=\; \big\lvert \mathbf{a}(\phi_a)^{\mathsf H}\, \mathbf{X}[r] \big\rvert,
  \qquad
  R[r,a] \;=\; 20 \log_{10} \big( S[r,a] + \epsilon \big),
  \label{eq:ra}
\end{equation}
on a grid whose numerical parameters are deployment-specific. Both
pipelines terminate in $R[r,a]$ on the same grid; we write
$R_{\mathrm{meas}}$ and $R_{\mathrm{sim}}$ for the two branches.

\textbf{Channel-conjugation convention.}\label{sec:design:signfix}
The receive-channel grid is complex-valued, and its phase convention
across virtual channels affects the angle FFT even when per-path
magnitudes are unchanged. A conjugated baseband convention preserves
per-channel magnitudes but reverses the sign of the phase slope across
the virtual array, which mirrors any non-zero angle response in the RA
map. We fix a single complex baseband convention shared by the
simulator's receive-channel grid and the measured-radar processing
branch before angle formation. The shared convention follows the
standard FMCW IF-demodulation expression $x[k] \propto e^{-j \pi k
\sin\theta}$ across the $k = 1, \ldots, M$ virtual channels; under this
convention, the simulator accumulates each path's contribution as
\begin{equation}
  \mathrm{RC}[r,k] \;\mathrel{+}=\; A_{p,r} \big( \cos\Phi_{p,r,k}
                                          - j\, \sin\Phi_{p,r,k} \big),
  \label{eq:signfix}
\end{equation}
so that a phase-slope inspection of the simulator's per-range-bin
virtual-array vector matches the phase-slope inspection of the measured
branch using the same convention.

\textbf{Mathematical validation.}
Under a convention flip, the resulting receive-channel grid is the
complex conjugate of the original for every validation pose, so the sign
rule performs no other operation than imposing the agreed convention. We
deliberately separate this \emph{convention validation} from the later
evaluation of residuals in the shared comparison domain.

\subsection{Per-Path Complex Contribution Records}
\label{sec:design:audit}
\label{sec:design:records}
For each surviving propagation path $p$, the simulator emits a
\emph{per-path complex contribution vector}
\begin{equation}
  \mathbf{c}_p \;=\; \big( c_{p,1}, c_{p,2}, \ldots, c_{p,M} \big)
                 \;\in\; \mathbb{C}^{M}
\end{equation}
over the $M$ virtual receive channels, together with metadata that
identifies the path's actor and material tag, bounce order, propagation
state, range bin $r_p$, and angle support. The contribution vector and
the corresponding update to the receive-channel grid are produced by
the same complex arithmetic, so the grid is reconstructible from the
contribution records by accumulation
\begin{equation}
  \mathrm{RC}[r_p, :] \;\mathrel{+}=\; \mathbf{c}_p,
  \label{eq:rc_accumulation}
\end{equation}
applied to every contribution record. This co-construction allows the
residual attribution framework to decompose the RA response at any
$(r, a)$ cell into the subset of paths that contribute to that cell, such
as paths belonging to a single named scene actor or a single material
tag. Residuals can then be attributed back to path subsets rather than
to aggregate cell values alone.

Let $A_p$ denote the linear amplitude assigned to path $p$ at the range
bin $r_p$ by the simulator's path-generation step. The contribution
vector then satisfies
\begin{equation}
  \sum_{k=1}^{M} \lvert c_{p,k} \rvert^{2} \;=\; A_p^{2} \cdot M,
  \label{eq:audit_invariant}
\end{equation}
which is the consistency condition between the path-level amplitude
assigned by the simulator and the per-channel contribution it writes. We
verify this consistency at machine precision for validation poses. The
condition acts as the audit invariant for path-level attribution and
makes the contribution records the primitive consumed by the residual
attribution framework.

Preserving complex contributions through the pipeline, rather than
collapsing each path to a magnitude before accumulation, retains the
phase information that the spatial FFT needs to form the angle
dimension.

\section{Residual Attribution for Calibration Decisions}
\label{sec:taxonomy}

\subsection{From Residual to Calibration Decision}
\label{sec:taxonomy:why}

Once both branches terminate in the same range-angle representation, the
question shifts from ``does the simulator match the radar?'' to ``given
a residual, which calibration action is appropriate?'' Without an answer
to the second question, the workflow faces a credit-assignment problem:
a material change might hide a residual that was actually caused by a
missing path, and a normalization change might mask a residual caused by
a sign-convention bug. Picking the wrong calibration action does not
just waste effort; it can produce a simulator that resembles the
measurement for the wrong reason.

The residual attribution framework turns each residual into a
calibration decision. For this calibration workflow, we use a decision
record of five discrepancy categories, and each category carries a
designated calibration action or limitation tag. These categories cover
the residuals encountered in the indoor office deployment studied in
this paper; we do not claim that they exhaust every mmWave simulator-measurement residual that a future workflow could surface. The record is extensible if a future workflow surfaces a residual that fits none of them. Material-RCS sweeps, in particular, are only admissible for Class~2 residuals, where a simulated path exists, but its response is weak. Within the record, the categories are mutually exclusive: every residual is assigned to exactly one category, and if a residual fits more than one, the assignment policy uses the lowest-numbered matching category.

\subsection{Discrepancy Categories and Calibration Actions}
\label{sec:taxonomy:classes}

The five categories below are calibration decision categories for the
workflow this paper describes, not a closed taxonomy of mmWave radar
discrepancies in general. Each category is defined operationally on the
simulated maximum within a match window around a geometry-supported
expected response region: a region in the range-angle representation
where the reconstructed scene geometry, radar pose, and expected object
extent jointly support the presence of a strong structural response.
Each category is paired with a designated calibration action or
limitation tag. Where a category's action refers to a physical mechanism
that the current platform has not yet enabled (Class~5 is the canonical
example), the action marks a mechanism-extension target rather than an
immediately available calibration parameter.

\noindent\textbf{Class 1 --- Missing path support.} The simulated RA has no
response near the geometry-supported expected response region, even
under a wider search window. The designated calibration actions are
scene-geometry repair, collision-proxy repair, actor-presence checks, or
trace-coverage changes; in short, any reason the simulator's ray set is
not reaching the expected structure.

\noindent\textbf{Class 2 --- Weak supported path.} The simulator has a
response within the tight match window, but its peak-normalized
amplitude falls below the strong-response threshold. The designated
calibration action is a material or RCS-level update on the responsible
actor's tagged material. This is the only category for which material or
RCS sweeps are admissible.

\noindent\textbf{Class 3 --- Unsupported anchor.} The simulator produces
a near-field, normalization-dominating, or otherwise unsupported peak
that is not explained by the geometry-supported expected response
region. The designated calibration action is system-response handling,
near-field gating, normalization review, or antenna-response modeling.

\noindent\textbf{Class 4 --- Shifted response.} The simulator has a
response near a geometry-supported expected response region, but the
peak is offset beyond the standard match window. The designated
calibration actions are pose correction, geometry placement, beamforming, 
and angle-convention review.

\noindent\textbf{Class 5 --- Missing physical mechanism.} A measured
structural response has no plausible simulated counterpart under the
current path basis. The designated action is to mark the case as a
mechanism-extension target, such as diffraction, diffuse scattering, or
floor-ceiling multipath. Class~5 is deliberately distinguished from
Class~1: Class~1 indicates missing path support under the current
mechanism set, whereas Class~5 indicates that the mechanism set itself
is incomplete. Table~\ref{tab:taxonomy} summarizes the categories, their
operational cues, and the calibration action or limitation tag assigned
to each category.

\begin{table}[!t]
  \centering
  \caption{Residual classes used for attribution and evaluation.}
  \label{tab:taxonomy}
  \begin{tabular}{@{}cll@{}}
    \toprule
    Class & Label & Cue \\
    \midrule
    1 & Missing support & No simulated response. \\
    2 & Weak support & Simulated response is weak. \\
    3 & Unsupported anchor & Unsupported peak dominates. \\
    4 & Shifted response & Response is range/angle shifted. \\
    5 & Missing mechanism & No path-basis counterpart. \\
    \bottomrule
  \end{tabular}
\end{table}
\subsection{Calibration-Domain Cross-Check}
\label{sec:design:domain}

The calibration domain is the fixed range-angle tensor on which the
residual attribution framework compares measurement and simulation. For
a given deployment, this tensor has shape
$N_{\mathrm{rng}} \times N_{\mathrm{ang}}$, where
$N_{\mathrm{rng}}$ is determined by the fast-time sampling, range-FFT
configuration, and indoor range of interest, and
$N_{\mathrm{ang}}$ is determined by the virtual-array angle transform
and the angular field of view used for comparison. These dimensions are
implementation choices that must remain fixed across all poses in an
experiment; they are not part of the platform abstraction and are not
retuned per pose.

The purpose of the calibration domain is to define a shared comparison
interface. The measured branch maps raw ADC samples into a range-angle
representation through the fixed signal-processing chain, and the
simulator branch is mapped into the same tensor shape before residuals
are interpreted. Residual attribution is performed only after both
branches share the same range and angle indexing, normalization
convention, and region of interest.

Thus, the calibration domain fixes the shared indexing and normalization
used to interpret residuals; it does not validate the physical
correctness of any individual simulated response. Concrete grid
dimensions, range coverage, angular coverage, thresholds, and matching
tolerances are deployment-specific parameters reported with the
experimental setup.

\subsection{From Attribution to Calibration Action}
\label{sec:taxonomy:remediation}

The five discrepancy categories each name an action or tag on the
physics-and-geometry side: scene repair, material or RCS update,
system-response review, geometry placement, or extension of the ray and
path mechanism set. The attribution framework separates material and RCS
adjustments from geometry and path-support recovery, as well as from
system-response limitations, so residuals are not routed to a generic
``material tuning'' default.

In this paper, the action bank is deliberately restricted to
interpretable simulator updates and limitation tags. If a residual is
caused by missing geometry, unsupported path structure, near-field
system response, or an unimplemented propagation mechanism, the
framework records that cause rather than hiding it behind a free-form
residual correction. This choice keeps the evaluation focused on what
the physics-only radar twin can explain, what can be repaired through
scene or simulator calibration, and what remains a simulator-model
limitation.

\section{Implementation}
\label{sec:impl}

\subsection{Hardware}
\label{sec:impl:hardware}

The deployment hardware instantiates the platform's generic FMCW radar
interface with concrete antenna, chirp, sampling, and capture
parameters. The radar front-end is a TI AWR2243BOOST evaluation module
paired with a DCA1000EVM capture card. The current chirp configuration
uses carrier frequency $f_c = 77$~GHz, bandwidth $B = 960$~MHz, chirp
slope $\mu = 75$~MHz/$\mu$s, $N_r = 64$ ADC samples per chirp, a
$5$~MHz sampling rate, and a $6~\mu$s ADC start time. The capture card
streams raw multi-channel I/Q samples to host memory for offline
processing.

\begin{figure}[t]
  \centering
  \includegraphics[
    width=\columnwidth,
    trim=0 0 0 0,
    clip
  ]{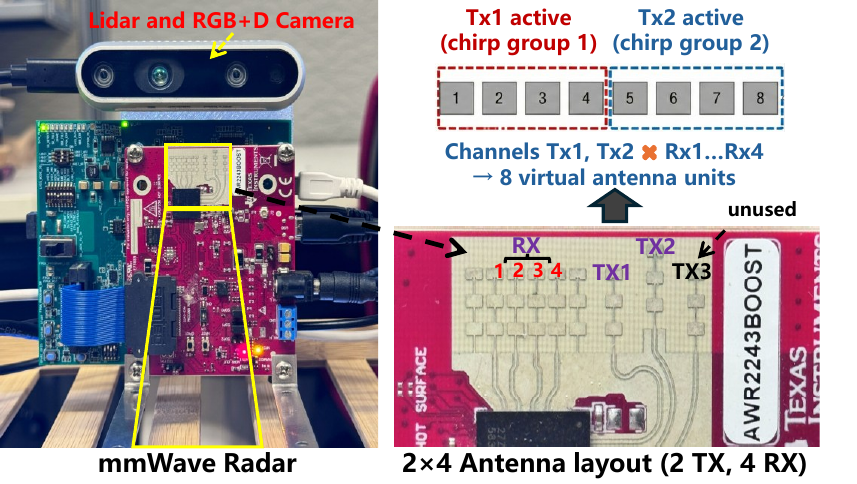}
  \caption{Hardware rig and TDM-MIMO antenna layout. The $N_T=2$,
  $N_R=4$ array forms an $M=8$ azimuth virtual array used by the shared
  range-angle processing pipeline.}
  \Description{Hardware figure showing the radar rig and the TDM-MIMO
  antenna layout used to form an eight-element azimuth virtual array.}
  \label{fig:hardware_rig}
\end{figure}

The radar operates in TDM-MIMO with $N_T = 2$ transmit and $N_R = 4$
receive antennas, producing a $M = N_T \times N_R = 8$-element virtual
array along the azimuth dimension. An Intel RealSense D435i depth-and-IMU
camera, rigidly mounted to the radar bracket, provides per-pose geometry
context and pose anchoring for the digital twin reconstruction; a laser
distance meter is used for calibration-pose verification. Fig.~\ref{fig:hardware_rig} shows the physical rig and the TDM-MIMO virtual-array layout used in the deployment.

We deliberately chose a single-radar monostatic configuration over a
distributed multi-radar configuration. Recent work has shown that
distributing three synchronized commodity AWR2243-class radars yields
polarimetric and bi-static SAR capabilities not available
monostatically~\cite{sun2026polysight}, at the cost of significant
cross-radar synchronization engineering (a hardware-trigger common
reference, a direct-path phase reference between radars, and a chirp
non-linearity compensation step).

\subsection{Scene Reconstruction and Material Library}
\label{sec:impl:scene}
The digital twin is built in Unreal Engine~5. We reconstruct the office
at two scales. At the macro scale, LiDAR-based simultaneous localization
and mapping captures room-level geometry, including walls, floors,
ceilings, and large furniture~\cite{xu2022fastlio2}. At the object
scale, a low-cost mobile object-capture workflow records overlapping
views of radar-relevant objects with an iPhone 15; Polycam Object Mode
then reconstructs textured mesh assets through mobile photogrammetry and
depth-assisted scanning. Prior studies report that, under well-lit
capture conditions, Polycam can achieve centimeter-level geometric
accuracy, which is sufficient for our object-placement and
mesh-alignment needs~\cite{VACATELLO2026e00480}. The two scales are
merged into a single UE5 level with a shared world coordinate system
aligned to the radar mounting fixture.

We regularize the reconstructed geometry before simulation. Mesh
decimation reduces triangle counts on object-scale meshes while
preserving silhouettes and large planar facets; the room-scale layout is
converted into cleaner surfaces to stabilize path extraction and bounce
counting; and simulation-friendly collision proxies are generated for
ray casting. The resulting geometry is reused unchanged across
calibration and what-if experiments.

Each reconstructed mesh is assigned one of seven paper-facing material
tags from the radar material library: Metal, Concrete, Wood, TableTop,
Carpet, Glass, or Plastic. These tags group reconstructed objects by the
radar-response behavior that the simulator needs to approximate, rather
than by a complete electromagnetic material characterization.

The Fresnel-dominant response used for Metal, Glass, Wood, and TableTop
is consistent with recent evidence that flat, smooth materials in the
high-incidence regime can behave electromagnetically smooth at the
implementation frequency~\cite{sun2026polysight}. For Carpet and
analogous soft-surface materials, we retain non-trivial diffuse and
backscatter coefficients, motivated by the Rayleigh roughness criterion:
a surface is electromagnetically rough when its RMS height exceeds
$\lambda/8$, and typical carpet and fabric surfaces can fall in this
rough regime~\cite{liu2026mmtexora}.

Table~\ref{tab:materials} reports the material-tag scalar defaults used
by the current deployment. BaseRCS sets the nominal specular response
strength, DiffBack sets the diffuse/backscatter response strength, and
BackExp, Mix, and NormExp control backscatter falloff,
specular/backscatter mixing, and backscatter normalization.

\begin{table}[!t]
  \centering
  \caption{Material-tag scalar defaults used by the current simulator deployment.}
  \label{tab:materials}
  \begin{tabular}{lccccc}
    \toprule
    Tag & BaseRCS & BackExp & Mix & DiffBack & NormExp \\
        & (dB)    & --      & --  & (dB)     & -- \\
    \midrule
    Metal     & $-3.0$  & 8.0  & 0.6 & $-99.0$ & 1.0 \\
    Concrete  & $-8.0$  & 2.0  & 0.0 & $-13.0$ & 1.0 \\
    Wood      & $-14.0$ & 2.5  & 0.0 & $-99.0$ & 1.0 \\
    TableTop  & $-22.0$ & 4.0  & 0.0 & $-99.0$ & 1.0 \\
    Carpet    & $-25.0$ & 0.5  & 0.0 & $-99.0$ & 1.0 \\
    Glass     & $-12.0$ & 10.0 & 0.0 & $-99.0$ & 1.0 \\
    Plastic   & $-16.0$ & 3.0  & 0.0 & $-99.0$ & 1.0 \\
    \bottomrule
  \end{tabular}
\end{table}

These values are deployment-fixed simulator parameters, not
independently measured material constants.

\subsection{UE Plugin Pipeline}
\label{sec:impl:plugin}

The mmRadarTwin UE plugin implements the simulation-side path tracing,
receive-channel accumulation, and per-path contribution-record export.
For each radar pose, the plugin traces the reconstructed scene,
accumulates the simulated receive-channel grid, and exports the
per-path complex contribution records used by the attribution
framework. The export also records the simulator configuration needed
to reproduce the pose-level run.

In the current deployment, the path tracer uses one fixed tracing
configuration across all poses, including bounce depth, path-power
cutoff, and ray budget. As a result, pose-to-pose differences reported
later are not caused by per-pose changes to tracing depth or sampling
budget.

\subsection{Real Data Pipeline}
\label{sec:impl:real}

The measured branch is mapped into the deployment calibration domain by
the same fixed signal-processing chain used for the simulated
receive-channel samples. Raw ADC samples from the capture board are
coherently averaged across $100$ frames per pose to reduce additive
measurement noise, Hann-windowed along the fast-time axis to suppress
sidelobe leakage, and passed through a fast-time range FFT.

The resulting complex range spectra on the $M=8$ virtual receive
channels are passed through the virtual-array angle FFT and converted to
log magnitude, yielding the measured range-angle representation
$R_{\mathrm{meas}}[r,a]$.

In this office deployment, the shared range-angle grid is instantiated
as $64 \times 64$. The range axis uses $64$ FFT bins under the current
ADC and chirp configuration, while the angle axis uses a $64$-bin
virtual-array angular output grid. These values instantiate the general
$N_{\mathrm{rng}} \times N_{\mathrm{ang}}$ calibration-domain interface:
they are fixed for all poses in this experiment but are not platform
constants. The measurement pipeline applies no per-pose tuning beyond
the fixed averaging count, window function, FFT lengths, and angular
output grid summarized in Table~\ref{tab:impl_domain}.
\begin{table}[!t]
  \centering
  \caption{Deployment range-angle grid and evaluation field of view.}
  \label{tab:impl_domain}
  \begin{tabular}{@{}lc@{}}
    \toprule
    Parameter & Current deployment value \\
    \midrule
    Grid shape $N_{\mathrm{rng}} \times N_{\mathrm{ang}}$
      & $64 \times 64$ \\
    Range-bin spacing
      & $0.15625$~m/bin \\
    Displayed range
      & $\approx 10$~m \\
    Angular output span
      & $-90^\circ$ to $+90^\circ$ \\
    Angular output bins
      & $64$ bins \\
    Main evaluation FOV
      & $-60^\circ$ to $+60^\circ$ \\
    \bottomrule
  \end{tabular}
\end{table}

\subsection{Dataset and Acquisition Coverage}
\label{sec:impl:dataset}

The acquisition campaign samples the office from multiple radar
locations, heights, and headings within the radar's usable field of
view. The final evaluation set contains 154 measured poses spanning
22 radar locations after excluding two duplicate measurement poses.
Each pose combines a radar location with a recorded mounting height and
heading. The poses are chosen to cover the major radar-visible portions
of the reconstructed scene, including wall-facing, cabinet-facing,
corridor, and oblique views. Experiments are organized by radar pose:
for each pose, the simulator query, the simulated receive-channel grid,
the derived simulated range-angle representation, and the measured
range-angle representation share a single pose record.

A fixed extrinsic transform between the radar mounting fixture and the
UE5 world frame aligns the measured and simulated branches at the pose
level and is applied consistently across the acquisition campaign. The
same deployment calibration grid is used for every pose, so differences
reported later are not produced by per-pose changes to the range or
angle discretization.

\begin{figure}[!t]
  \centering
  \includegraphics[
    width=\columnwidth,
    trim=40 50 40 50,
    clip
  ]{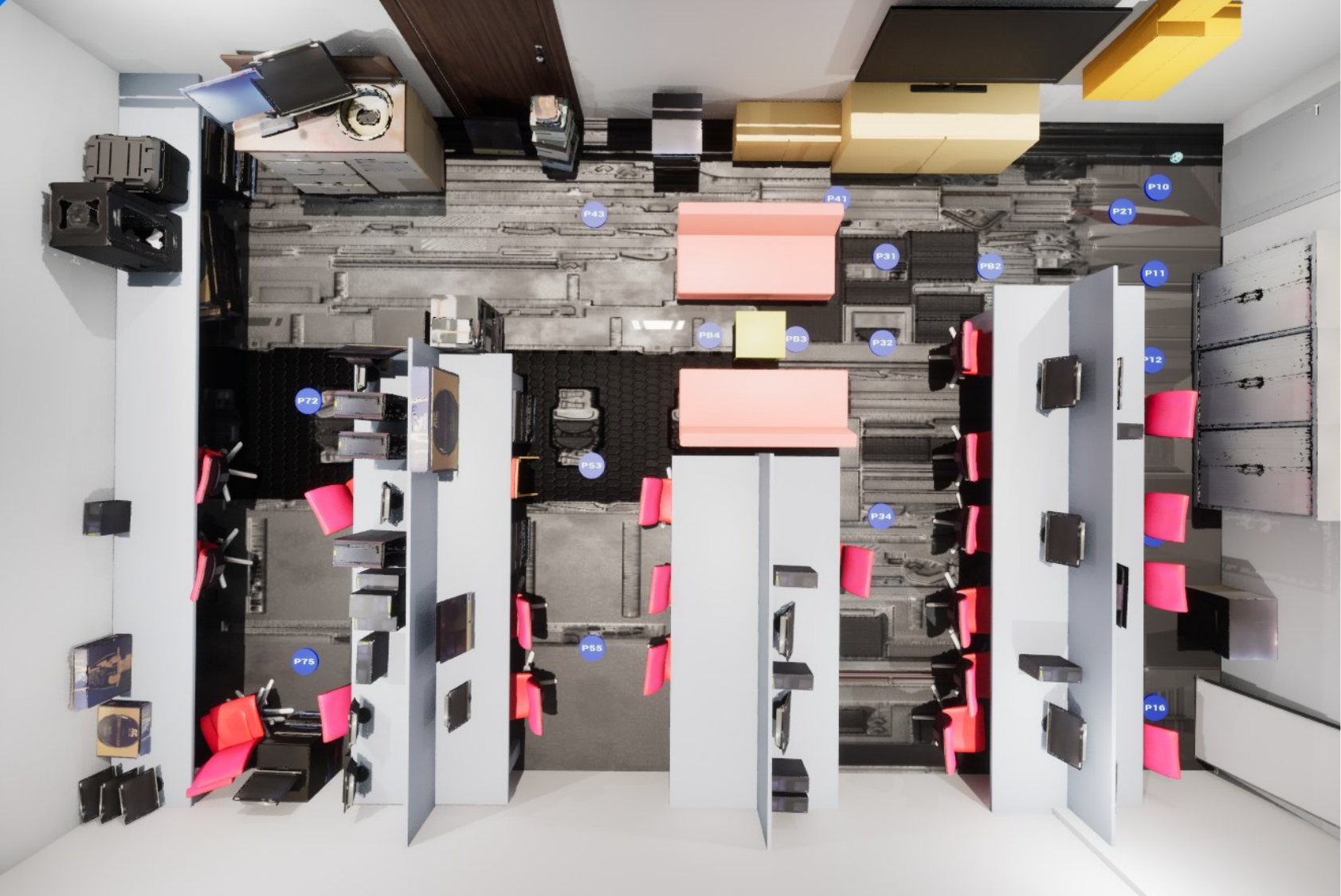}
  \caption{Acquisition coverage in the reconstructed indoor scene.
    Blue markers show the 22 sampled radar locations across desk, cabinet,
    corridor, and oblique viewpoints. Each evaluation pose combines one
    location with a recorded height and heading; the final evaluation set
    contains 154 measured poses after excluding duplicate measurement
    poses.}
  \Description{Top-down reconstructed office scene with radar-pose
  markers across desks, cabinets, corridor regions, and oblique
  viewpoints.}
  \label{fig:scene_radars}
\end{figure}

Figure~\ref{fig:scene_radars} visualizes the reconstructed indoor scene
and the sampled radar positions used by the acquisition campaign. The
top-down view documents the spatial coverage of the room-scale
evaluation area, including desk regions, cabinet-facing views,
corridor-facing views, and oblique viewpoints. The pose icons indicate
radar locations in the reconstructed scene; headings and heights are
stored in the pose records rather than fully represented in the
top-down visualization. Diagnostic cases used later in the paper are
treated as analysis cases within this acquisition coverage, not as a
separate dataset contribution.

\section{Evaluation}
\label{sec:eval}

\subsection{Evaluation Goal and Metrics}
\label{sec:eval:framework}

We evaluate mmRadarTwin as a physics-only, signal-level radar twin
rather than as an image-translation or image-refinement system. The
evaluation answers three questions. First, does the simulator place
strong responses in geometry-supported regions where the measured radar
also observes strong responses? Second, when the simulator and
measurement disagree, how large are the under-prediction,
over-prediction, supported-response amplitude error, and weak-background
perturbation? Third, which residuals remain outside the current
simulator model?

\noindent\textbf{Measurement reference.}
We use the measured radar response as the measurement reference in the
calibration domain. The measurement is not an absolute annotation of
scatterer locations: it includes hardware noise floor, antenna response,
multipath, diffuse scattering, and near-field effects that the current
simulator does not fully model.

\noindent\textbf{Normalization, thresholds, and field of view.}
Unless otherwise stated, both measured and simulated range-angle maps
are independently peak-normalized to $0$~dB for each pose and clipped to
$[-50,0]$~dB. This convention evaluates relative placement and residual
structure, not absolute received-power calibration. Strong, medium, and
weak cells are defined by thresholds of $-15$~dB, $-20$~dB, and
$-25$~dB, respectively. Cells within the hardware-near range
($r \leq 0.16$~m) are excluded. The main evaluation uses the central
usable field of view, $[-60^\circ,60^\circ]$ in azimuth. Responses in
the outer angular region, $[-90^\circ,-60^\circ) \cup
(60^\circ,90^\circ]$, are treated as appendix diagnostics and are not
routed to the main-paper C1-C5 counts. For range-stratified reporting,
we divide the map into near ($0.16$--$2.0$~m), mid ($2.0$--$4.0$~m),
and far ($>4.0$~m) bands.

\noindent\textbf{Matched-response metrics.}
For geometry-supported expected response regions, we report whether a
corresponding simulator response appears in the same range-angle
neighborhood and how close its magnitude is to the measured response.
The primary quantities are matched-region recall, Top-$K$ peak
agreement, and residual dB error over both-strong matched regions. A
region-level match uses a tolerance of $\pm0.20$~m or one range bin,
whichever is larger, and $\pm5^\circ$ or one angle bin, whichever is
larger. Peak agreement uses a wider peak-match tolerance of
$0.35$~m and $10^\circ$.

\noindent\textbf{Mismatch ledger.}
We categorize cell-level residuals into four diagnostic axes. Q1 denotes
cells where the measurement is strong but the simulator is weak,
capturing under-predicted measured returns. Q2 denotes cells where the
simulator is strong but the measurement is weak, capturing simulator
false positives. Q3 denotes cells where both maps are strong inside
expected regions, where magnitudes can be compared directly. Q4 denotes
cells where both maps are weak. Q4 is reported only as a
background-perturbation sanity check, not as a success metric. Q1 energy
is the linear-power mass in Q1 cells normalized by the measurement
energy for that pose; Q2 energy is the simulator linear-power mass in
Q2 cells normalized by the simulator energy for that pose.
\subsection{Scene-Grounded Real-to-Sim Comparison}
\label{sec:eval:scene_grounding}

\begin{figure*}[t]
  \centering
  \includegraphics[
    width=0.96\textwidth
  ]{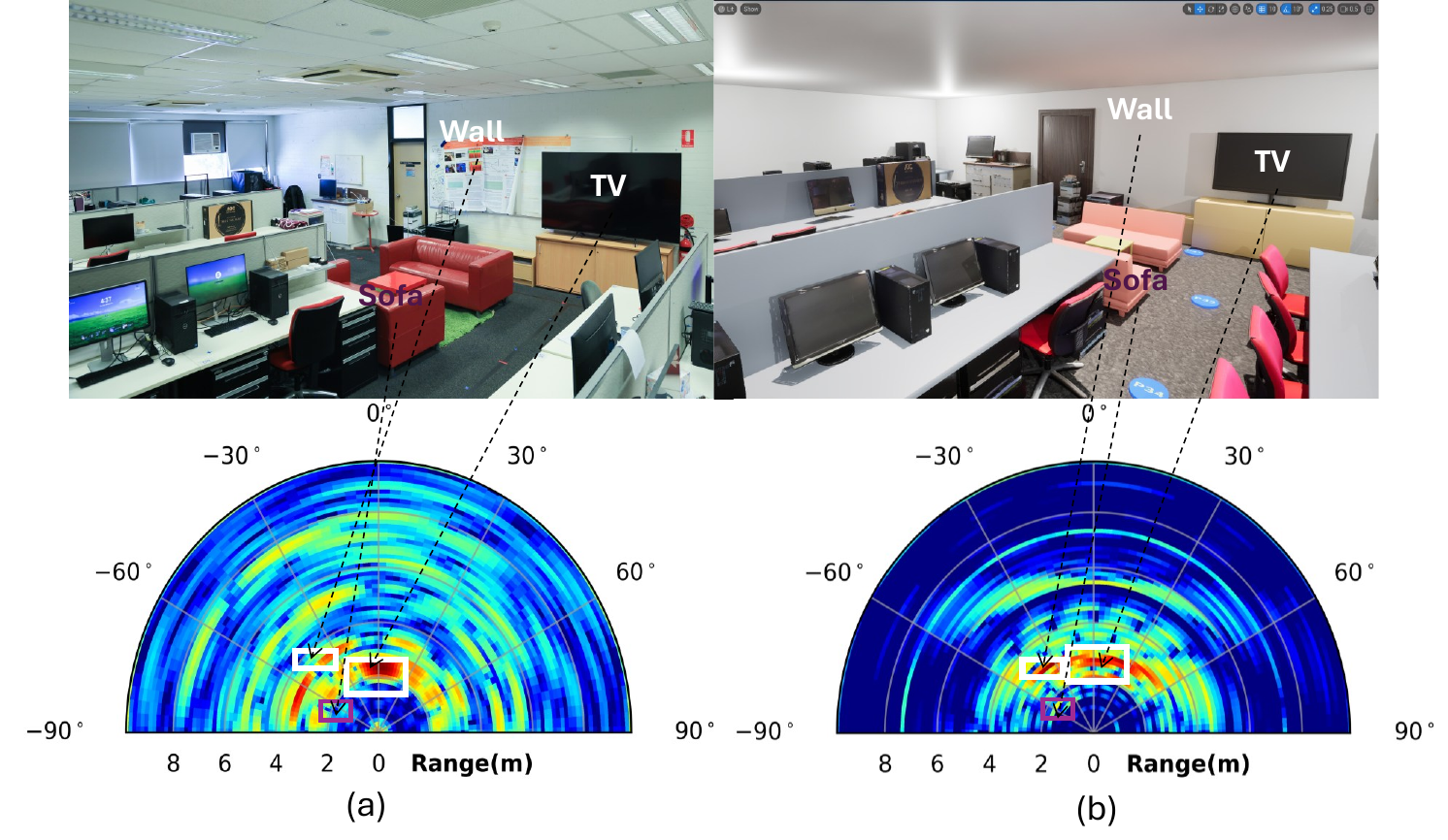}
  \caption{Scene-grounded real-to-sim comparison for one representative
  pose. Column~(a) shows the real scene and the measured radar response;
  Column~(b) shows the reconstructed UE scene and the simulator response
  under the same radar pose and range-angle interface. White boxes mark
  matched response regions, such as the wall and TV responses. The
  magenta box marks a mismatch region associated with the sofa, where
  the simulator response is stronger than the measurement. The marked
  regions are scene-grounded response regions, not absolute scatterer
  annotations.}
  \Description{Real scene, reconstructed UE scene, measured radar
  response, and simulator response for one pose, with matched and
  mismatched response regions marked.}
  \label{fig:eval_scene_grounded}
\end{figure*}

Before aggregating metrics across poses, we first ground the comparison
in the reconstructed scene. Figure~\ref{fig:eval_scene_grounded} shows one representative
scene-grounded comparison case. Column~(a) pairs the real scene with the
measured radar response. Column~(b) pairs the reconstructed UE scene
with the corresponding simulator response under the same radar pose.
This layout makes the evaluation target explicit: mmRadarTwin is not
judged by generic heatmap similarity but by whether scene-supported
objects and structures produce corresponding responses in the shared
range-angle domain.

The marked regions illustrate the two outcomes used throughout the
evaluation. White boxes indicate matched response regions, where the
measurement and simulator both contain a response supported by the
reconstructed scene geometry and radar pose. In this example, the wall
and TV regions are matched. The magenta box indicates a mismatch region:
the sofa-associated response is stronger in the simulator than in the
measurement, illustrating the type of over-predicted response counted by
the Q2 diagnostic axis. These boxes are not absolute scatterer
annotations. They are scene-grounded response regions used to organize
the comparison and to connect radar residuals back to objects and
structures in the reconstructed scene.

\subsection{Representative Match and Mismatch Examples}
\label{sec:eval:examples}

\begin{figure*}[t]
  \centering
  \includegraphics[width=0.92\textwidth]{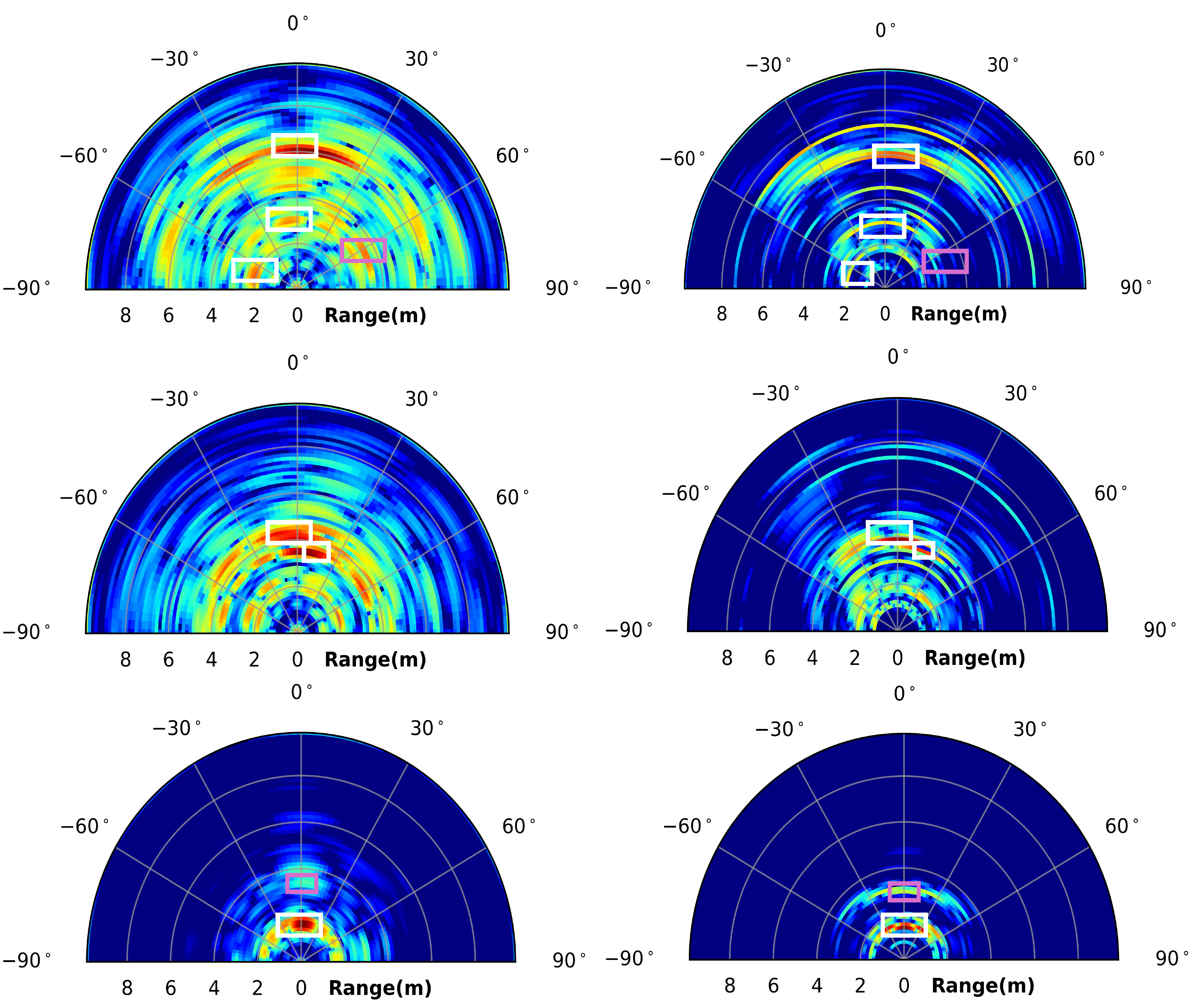}
  \caption{Representative match and mismatch examples. Each row compares
  the measured radar response (left) with the simulator response (right)
  for the same pose. White boxes mark matched response regions supported
  by reconstructed scene geometry and radar pose. Magenta boxes mark
  mismatch regions where one branch produces a stronger response than
  the other. These examples are qualitative; aggregate statistics over
  the central usable field of view are reported in
  Table~\ref{tab:eval_ledger}.}
  \Description{Three representative poses comparing measured and
  simulated radar responses, with matched and mismatched response
  regions marked.}
  \label{fig:eval_matched_examples}
\end{figure*}

Figure~\ref{fig:eval_matched_examples} shows additional representative
poses in the shared range-angle domain. Each row compares the measured
radar response with the simulator response for the same radar pose.
White boxes mark matched response regions supported by reconstructed
scene geometry and radar pose. Magenta boxes mark mismatch regions where
one branch produces a stronger response than the other.

These examples are qualitative. They show that the physics-only
simulator can reproduce a subset of scene-supported strong responses,
while also exposing residuals that remain unmatched or over-predicted.
The figure should therefore be read together with
Table~\ref{tab:eval_ledger}, which reports aggregate statistics over the
full evaluation set.

\subsection{Quantifying Match and Mismatch}
\label{sec:eval:ledger}

Table~\ref{tab:eval_ledger} summarizes the evaluation over $154$
measured poses after excluding two duplicate measurement poses. The
table is organized around the evidence needed to judge the simulated
radar maps: response placement, global peak agreement, matched-region
amplitude error, mismatch mass, background perturbation, and residual
routing. The first three rows evaluate where the simulator agrees with
the measurement; the remaining rows quantify where it under-predicts,
over-predicts, perturbs weak-background regions, or routes residuals to
the C1-C5 attribution classes.

\begin{table}[!t]
  \centering
  \caption{Evaluation summary in the central usable field of view
  ($[-60^\circ,60^\circ]$).}
  \label{tab:eval_ledger}
  \begin{tabular*}{\columnwidth}{@{\extracolsep{\fill}}lll@{}}
    \toprule
    Target & Metric & Result \\
    \midrule
    Placement & Recall & $0.708$; $0.828/0.607/0.384$ \\
    Peaks & Top-$1$/$3$/$5$ & $0.260/0.242/0.261$ \\
    Matched amp. & Q3 median/IQR & $1.01/5.80$~dB \\
    Under-pred. & Q1 energy & $0.518$ \\
    Over-pred. & Q2 energy & $0.311$ \\
    Background & Q4 MAE & $7.20$~dB \\
    Routing & C1/C2/C3/C4/C5 & $3/26/16/33/22\%$ \\
    \bottomrule
  \end{tabular*}
\end{table}

The strongest result is region-level placement. The simulator recalls
$70.8\%$ of measurement-active geometry-supported response regions in
the central usable field of view. Recall is highest in the near range
($82.8\%$), decreases in the mid range ($60.7\%$), and is lowest in the
far range ($38.4\%$), where diffuse wall response, multipath, and
environment-specific ghost structure become more prominent.

The lower Top-$K$ agreement should be read as a strict global-peak
metric. In the measured data, dominant peaks can shift between nearby
objects and far wall responses across headings, reflecting
pose-dependent visibility, material response, multipath, and
environment-specific ghost structure. This behavior makes global peak
ordering harder to reproduce than scene-grounded region-level placement.

The mismatch metrics show the current limitations. Q1 and Q2 remain
substantial, meaning the simulator still under-predicts some measured
returns and over-predicts some responses that are weak in the
measurement. The residual routing row explains why these errors should
not be collapsed into one material-tuning problem: C2 is the direct
material/RCS-update category, while C1, C3, C4, and C5 point to path
support, system response, geometry or beamforming shift, and missing
physical mechanisms.

\subsection{Residual Attribution and Model Limits}
\label{sec:eval:limits}

The residual ledger separates calibration targets from simulator-model
limits. Some residuals point to repairable causes: missing path support,
weak supported paths, shifted responses, or system-response effects that
suggest scene repair, material/RCS update, pose correction, beamforming
review, or normalization review. Other residuals expose limits of the
current path basis, including diffuse scattering, multipath or ghost
responses, fine geometry missing from the scene model, and hardware
noise-floor effects.

\begin{figure}[t]
  \centering
  \includegraphics[width=\columnwidth]{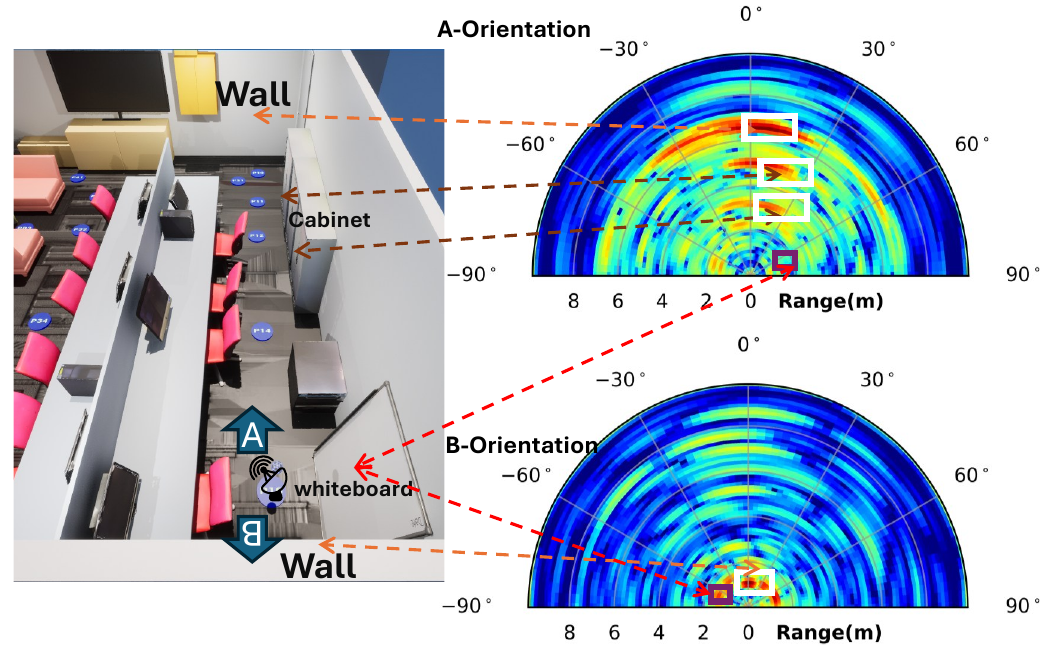}
  \caption{Pose-dependent peak dominance in the measured radar response.
Left: UE scene context for one radar location, with two opposite
measurement headings marked as A-orientation and B-orientation. Right:
measured radar range-angle responses collected from the same location
under the two headings. The marked regions show that the dominant
measured peaks can shift between nearby structures and wall responses
when the radar heading changes. This heading-dependent behavior makes
global Top-$K$ peak agreement a strict auxiliary metric and motivates a
scene-grounded region-level evaluation.}
  \label{fig:topk_limitation}
\end{figure}

Figure~\ref{fig:topk_limitation} illustrates why global peak agreement
is a strict auxiliary metric. Two opposite headings at the same radar
location can produce different dominant measured peaks: nearby
structures may dominate in one heading, while wall responses dominate in
the opposite heading. This pose-dependent peak dominance makes global
Top-$K$ agreement harder than scene-grounded region-level placement and
helps explain why Table~\ref{tab:eval_ledger} treats Top-$K$ as an
auxiliary metric rather than the main success criterion.

This distinction matters most for false positives. A simulator-strong
and measurement-weak response is not automatically a material error. It
may indicate unsupported path geometry, an over-visible surface, an
angular-response mismatch, or a path that should be suppressed by a
physical mechanism not yet modeled. For this reason, Q2 residuals are
reported as calibration and limitation evidence rather than as a solved
metric.

The paper-taxonomy remap confirms that residuals are not dominated by a
single missing-mechanism category. Within the central usable field of
view, the largest residual class is C4 shifted response ($33\%$),
followed by C2 weak supported path ($26\%$), C5 missing physical
mechanism ($22\%$), C3 unsupported anchor ($16\%$), and C1 missing path
support ($3\%$). This distribution is useful because it indicates which
calibration action is appropriate. Only C2 is a direct material/RCS
update case. C1 and C4 point to geometry, path-support, pose, or
beamforming correction. C3 points to system-response or normalization
limits. C5 marks propagation mechanisms that the current simulator does
not yet implement.

Weak-background regions are treated separately. The measured and
simulated maps can both be weak while still differing in low-level
texture because of hardware noise, diffuse scattering, and unmodeled
multipath. We therefore report Q4 as a background perturbation metric
and do not use it to claim success.

\subsection{Summary of Evidence}
\label{sec:eval:summary}

The evaluation supports a scoped claim. In one office deployment,
mmRadarTwin recovers a substantial subset of geometry-supported measured
responses in the shared range-angle domain: $70.8\%$ matched-region
recall overall, with stronger recovery in the near range and weaker
recovery in the far range. Within both-strong expected regions, the
median simulator-measurement residual is $1.01$~dB, indicating that
supported strong responses can be close in magnitude once both branches
place energy in the same region.

The remaining evidence is equally important. Q1 under-predictions and
Q2 false positives remain substantial, with normalized energy masses of
$0.518$ and $0.311$, respectively. Residual routing shows that these
errors are distributed across shifted responses, weakly supported paths,
unsupported anchors, and missing mechanisms rather than being explained
by one error source. These results support mmRadarTwin as an
interpretable radar-twin workflow for constructing, comparing, and
diagnosing indoor radar simulations. They do not imply complete
radar-map reconstruction, cross-room generalization, or full agreement
between the simulator and measured radar response.

\section{Related Work and Discussion}
\label{sec:related}

Ray tracing and wireless digital twins provide the closest simulation
context for mmRadarTwin. Sionna~RT and differentiable ray-tracing
extensions expose channel impulse responses, path gains, and
gradient-based calibration interfaces for wireless
systems~\cite{hoydis2023sionnart,Hoydis2024DiffRT,Ruah2024CalibRT}.
Other RT-based digital-twin systems target site-specific communication
deployment, RF sandboxes, or vehicular and urban channel
analysis~\cite{Zhang2023WiSegRT,WirelessDTThesis,Salehi2024Multiverse,
Li2022UrbanRT,He2023AutoRadar,Wang2024AutoRadar}. These systems are
highly relevant, but their output layer is usually a channel, path, or
path-loss quantity. mmRadarTwin instead targets the complex
receive-channel representation consumed by an indoor radar processing
chain and preserves per-path contribution records so that residuals in
the range-angle domain can be attributed to scene, material, system, or
mechanism causes.

A second line of work synthesizes or reconstructs mmWave sensing data
with hybrid physics-learning pipelines. RF~Genesis combines custom ray
tracing with diffusion models to synthesize spectrogram- and DRAI-level
data for sensing classifiers~\cite{chen2023rfgenesis}. Stereo-Fi
co-trains a physics-based model with a learned generative model for
free-form 3D reconstruction from RF measurements~\cite{han2026stereofi}.
Neural channel and RF-field representations further show how implicit
models can represent wireless propagation fields or radar-like
signals~\cite{Zhao2023NeRF2,Zhao2024NeRF2Channel,Lu2024NeWRF,
Zhang2025RF3DGS,Jiang2025LearnableDT,Orekondy2023WINERT,
Liu2025DynamicIndoorMIMO,Borts2024RadarFields}. mmRadarTwin is
complementary to this line: it does not use a learned image translator
as the main method, but asks how far a physics-only simulator can go
when its residuals remain signal-level and path-attributed.

A third line characterizes the material and hardware regime of mmWave
sensing. POLySight shows that distributed polarimetric measurements on
commodity mmWave radars can classify materials by exploiting Fresnel
polarization near the Brewster angle~\cite{sun2026polysight}. mmTexora
studies rough-surface material response through ambient-vibration-induced
temporal phase variation and the Rayleigh criterion~\cite{liu2026mmtexora}.
These works motivate our distinction between smooth, Fresnel-dominant
tags and rough or diffuse surfaces, but mmRadarTwin uses material tags
as simulator parameters rather than as independently measured material
constants.

The present system has four main limitations. First, far-range wall
responses, diffuse scattering, and multipath or ghost returns remain
difficult for the current path basis. Second, false-positive simulator
responses and shifted responses often require geometry, pose,
beamforming, or system-response fixes rather than simple material/RCS
updates. Third, the current evaluation uses scene-grounded response
regions and range-angle metrics, but it does not yet connect the
simulator to a downstream radar-perception task. Finally, the current
implementation is validated in one office environment; rebuilding the
platform for additional rooms is necessary to test how well the workflow
generalizes across layouts and material distributions.

\section{Conclusion}
\label{sec:conclusion}

mmRadarTwin is a signal-level and path-attributed digital-twin platform
for indoor mmWave radar. It connects a measured radar branch and an
Unreal Engine scene-simulation branch through a shared range-angle
comparison domain, while preserving per-path complex contribution
records that link simulated returns to actors, material tags,
propagation events, and output bins. This design turns
simulator-measurement residuals into diagnostic evidence: some
mismatches point to material/RCS updates, while others point to geometry,
path support, pose, beamforming, system response, or missing physical
mechanisms. Our office deployment shows that the physics-only simulator
recalls a substantial fraction of geometry-supported expected response
regions in the central field of view while also exposing false positives,
shifted responses, diffuse scattering, multipath, fine geometry, and
hardware noise-floor effects as current model limits. These results
support mmRadarTwin as an interpretable radar-twin workflow for indoor
sensing research, not as a claim that the simulator fully matches every
measured radar response.


\bibliographystyle{ACM-Reference-Format}
\bibliography{mmRadarTwin_icml_clean}

\end{document}